\documentclass[11pt,a4paper]{article}
\usepackage[hyperref]{acl2020}
\usepackage{times}
\usepackage{latexsym}
\usepackage{color}
\usepackage{multirow}
\usepackage{url}
\usepackage{graphicx,subcaption}
\usepackage{amsmath,amssymb}
\usepackage{xspace}
\usepackage{xcolor}

\aclfinalcopy %
\def\aclpaperid{1990} %

\newcommand{\la}{\textsc{LSTMavg}\xspace}
\newcommand{\wa}{\textsc{Wordavg}\xspace}
\newcommand{\sro}{\textsc{LSTMGaussian}\xspace}
\newcommand{\ws}{\textsc{Wordsum}\xspace}
\newcommand{\wlo}{WLO\xspace}

\newcommand{\scavg}{\textsc{avg}\xspace}
\newcommand{\scsum}{\textsc{sum}\xspace}
\title{Learning Probabilistic Sentence Representations from Paraphrases}

\author{Mingda Chen\qquad Kevin Gimpel\\
Toyota Technological Institute at Chicago, Chicago, IL, 60637, USA\\
  {\tt \{mchen,kgimpel\}@ttic.edu}\\}

\date{}

\begin{document}
\maketitle
\begin{abstract}

Probabilistic word embeddings have shown effectiveness in capturing notions of generality and entailment, but there is very little work on doing the analogous type of investigation for sentences. In this paper we define probabilistic models that produce distributions for sentences. Our best-performing model treats each word as a linear transformation operator applied to a multivariate Gaussian distribution.
We train our models on paraphrases and demonstrate that they naturally capture sentence specificity.
While our proposed model
achieves the best performance overall, we also show that specificity is represented by simpler architectures via the norm of the sentence vectors.
Qualitative analysis shows that our probabilistic model captures sentential entailment and provides ways to analyze the specificity and preciseness of individual words.

\end{abstract}

\section{Introduction}

Probabilistic word embeddings have been shown to be useful for capturing notions of generality and entailment~\cite{vilnis2014word,ath2017multimodal,athiwaratkun-etal-2018-probabilistic}. In particular, researchers have found that the entropy of a word roughly encodes its generality, even though there is no training signal explicitly targeting this effect.
For example, hypernyms tend to have larger variance than their corresponding hyponyms~\cite{vilnis2014word}. However, there is very little work on doing the analogous type of investigation for sentences.

In this paper, we define probabilistic models that produce distributions for sentences. In particular, we choose a simple and interpretable probabilistic model that
treats each word as an operator that translates and scales a Gaussian random variable representing the sentence.
Our models are able to capture sentence specificity as measured by the annotated datasets of \citet{li2015fast} and \citet{ko2018domain} by training solely on noisy paraphrase pairs. 
While our ``word-operator'' model yields the strongest performance, we also show that specificity is represented by simpler architectures via
the norm of the sentence vectors.
Qualitative analysis shows that our models represent sentences in ways
that correspond to the entailment relationship
and
that individual word parameters can be analyzed to find words with varied and precise meanings.

\section{Proposed Methods}
We propose a model that uses ideas from flow-based variational autoencoders (VAEs)~\cite{rezende15norm-flow,kingma2016improved} by treating each word as an ``operator''. 
Intuitively, we assume there is a random variable
$z$ associated with each sentence $s=\{w_1,w_2,\cdots,w_n\}$.
The random variable initially follows a standard multivariate Gaussian distribution.
Then, each word in the sentence transforms the random variable sequentially, leading to a random variable that encodes its semantic information.

Our word linear operator model (\wlo)
has two types of parameters for each word $w_i$: a scaling factor $A_i\in\mathbb{R}^k$ and a translation factor $B_i\in\mathbb{R}^k$.
The word operators produce a sequence of random variables $z_0,z_1,\cdots,z_n$ with $z_0\sim\mathcal{N}(0,I_k)$, where $I_k$ is a $k\times k$ identity matrix, and the operations are defined as
\begin{equation}
    z_i = A_i(z_{i-1}+B_i)
\end{equation}

\noindent The means and variances for each random variable are computed as follows:
\begin{align}
    \mu_i&=A_i(\mu_{i-1}+B_i)\\
    \Sigma_i&=A_i\Sigma_{i-1}A_i^\top
\end{align}

\noindent
For computational efficiency, we only consider diagonal covariance matrices, so the equations above can be further simplified.

\section{Learning}
Following \citet{para-nmt-acl-18}, all of our models are trained with a margin-based loss on paraphrase pairs $(s_1, s_2)$:
\begin{align}
    &\max(0,\delta-d(s_1,s_2)+d(s_1,n_1)) + \nonumber\\
    &\max(0,\delta-d(s_1,s_2)+d(s_2,n_2)) \nonumber
\end{align}
\noindent where $\delta$ is the margin and $d$ is a similarity function that takes a pair of sentences %
and outputs a scalar denoting their similarity. 
The similarity function is maximized over a subset of examples (typically, the mini-batch) to choose negative examples $n_1$ and $n_2$. %
When doing so, we use ``mega-batching'' \citep{para-nmt-acl-18} and fix the mega-batch size at 20. For deterministic models, $d$ is cosine similarity, while for probabilistic models, we use the expected inner product of Gaussians.

\subsection{Expected Inner Product of Gaussians}%
Let $\mu_1$, $\mu_2$ be mean vectors and $\Sigma_1$, $\Sigma_2$ be the variances %
predicted by models for a pair of input sentences. For the choice of $d$,  
following \citet{vilnis2014word}, we use the expected inner product of Gaussian distributions: %
\begin{equation}
\begin{aligned}
    &\int_{x\in\mathbb{R}^k}\mathcal{N}(x;\mu_1,\Sigma_1)\mathcal{N}(x;\mu_2,\Sigma_2)dx\\
    &=\log\mathcal{N}{(0;\mu_1-\mu_2,\Sigma_1+\Sigma_2)}\\
    &=-\frac{1}{2}\log\det{(\Sigma_1+\Sigma_2)}-\frac{d}{2}\log(2\pi)\\
    &\phantom{ = }-\frac{1}{2}(\mu_1-\mu_2)^\top(\Sigma_1+\Sigma_2)^{-1}(\mu_1-\mu_2)
\end{aligned}
\end{equation}
\noindent 
For diagonal matrices $\Sigma_1$ and $\Sigma_2$, the equation above can be computed analytically. 

\subsection{Regularization}

To avoid the mean or variance of the Gaussian distributions from becoming unbounded during training, resulting in degenerate solutions, we impose prior constraints on the operators introduced above. %
We force the transformed distribution after each operator to be relatively close to $\mathcal{N}(0,I_k)$, which can be thought of as our ``prior'' knowledge of the operator. Then our training additionally minimizes 
\begin{equation}
\begin{aligned}
    &\lambda\!\!\!\!\sum_{s\in\{s_1,s_2,n_1,n_2\}}\sum_{w\in s}\mathit{KL}(\mathcal{N}(\mu(w),\Sigma(w))\Vert\mathcal{N}(0,I))\nonumber
\end{aligned}
\end{equation}
\noindent where $\lambda$ is a hyperparameter tuned based on the performance on the 2017 semantic textual similarity (STS; \citealp{cer2017semeval}) data. 
We found prior regularization very important, 
as will be shown in our results. 
For fair comparison, we also add L2 regularization to the baseline models.

\section{Experiments}
\subsection{Baseline Methods}
We consider two baselines that have shown strong results on sentence similarity tasks \citep{para-nmt-acl-18}. The first, word averaging (\wa), simply averages the word embeddings in the sentence. The second, long short-term
memory~(LSTM;~\citealp{hochreiter1997long}) averaging (\la), uses an LSTM to encode the sentence and averages the hidden vectors. Inspired by sentence VAEs~\cite{bowman16gen}, we consider an LSTM based probabilistic baseline (\sro) which builds upon \la and uses separate linear transformations on the averaged hidden states to produce the mean and variance of a Gaussian distribution. 

We also benchmark several pretrained models, including GloVe~\cite{glove}, Skip-thought~\cite{Kiros2015skipthought}, InferSent~\cite{infersent}, BERT~\cite{devlin-etal-2019-bert}, and ELMo~\cite{peters-etal-2018-deep}. When using GloVe, we either sum embeddings (GloVe \scsum) or average them (GloVe \scavg) to produce a sentence vector. Similarly, for ELMo, we either sum the outputs from the last layer (ELMo \scsum) or average them (ELMo \scavg). For BERT, we take the representation for the ``[CLS]'' token.

\begin{table}[t]
    \small
    \centering
    \begin{tabular}{l|c|c|c|c}
         Domain & News & Twitter & Yelp & Movie \\\hline
         Number of instances & 900  & 984 & 845 & 920
    \end{tabular}
    \caption{Sizes of %
    test sets for sentence specificity.}
    \label{tab:data}
\end{table}

\subsection{Datasets}
We use the preprocessed version of ParaNMT-50M~\cite{para-nmt-acl-18} as our training set, which consists of 5 million paraphrase pairs.

For evaluating sentence specificity, we use human-annotated test sets from four domains, including news, Twitter, Yelp reviews, and movie reviews, from \citet{li2015fast} and \citet{ko2018domain}. 
For the news dataset, labels are either ``general'' or ``specific'' and there is additionally a training set. For the other datasets, labels are real values indicating specificity. 
Statistics for these datasets are shown in Table~\ref{tab:data}.

For analysis we also use the semantic textual similarity (STS) 
benchmark test set~\cite{cer2017semeval} and the Stanford Natural Language Inference (SNLI) dataset~\citep{bomwn2015large}.

\subsection{Specificity Prediction Setup}
For predicting specificity in the news domain, we threshold the predictions either based on the entropy of Gaussian distributions produced from probabilistic models or based on the norm of vectors produced by deterministic models, which includes all of the pretrained models. The threshold is tuned based on the training set but no other training or tuning is done for this task with any of our models. For prediction in other domains, we simply compute the Spearman correlations between the entropy/norm and the labels.

Intuitively, when sentences are longer, they tend to be more specific. %
So, we report baselines (``Length'') that predict specificity solely based on length, by thresholding the sentence length for news (choosing the threshold using the training set) or simply returning the length for the others. The latter results are reported from \citet{ko2018domain}. %
We also consider baselines that average or sum ranks of word frequencies within a sentence (``Word Freq.~\scavg'' and ``Word Freq.~\scsum'').

\section{Results}
 
\subsection{Sentence Specificity}

Table~\ref{sent-spe-gen-res} shows results %
on sentence specificity tasks. We compare to the best-performing models reported by
\citet{li2015fast} and \citet{ko2018domain}.  %
Their models are specifically designed for predicting sentence specificity and they both use labeled training data from the news domain.

\begin{table}[t]
\setlength{\tabcolsep}{5pt}
\centering
\small
\begin{tabular}{l|rrrr}
 & \multicolumn{1}{|c}{News} & \multicolumn{1}{c}{Twitter} & \multicolumn{1}{c}{Yelp} & \multicolumn{1}{c}{Movie} \\\hline
Majority baseline & 54.6 & \multicolumn{1}{c}{-} & \multicolumn{1}{c}{-} & \multicolumn{1}{c}{-} \\
Length & 73.4 & 44.5  & 67.6 & 58.1 \\
Word Freq.~\scsum & 55.5 & 10.1 & 54.6 & 22.1 \\ 
Word Freq.~\scavg & 61.5 & 0.0 & 28.5 & 0.0 \\\hline
\multicolumn{5}{c}{Prior work trained on labeled sentence specificity data}
\\\hline
\citet{li2015fast} & 81.6 & 55.3  & 63.3 & 57.5 \\
\citet{ko2018domain} & \multicolumn{1}{|c}{-} & 67.9 & 75.0 & 70.6 \\\hline
\multicolumn{5}{c}{Sentence embeddings from pretrained models} \\
\hline
GloVe \scsum & 70.4 & 32.2 & 62.8 & 49.0 \\
GloVe \scavg & 54.6 & -49.6 & -59.0 & -38.2 \\
InferSent & 75.0 & 60.5 & 76.6 & 61.2 \\
Skip-thought & 57.7 & 2.9 & 14.1 & 27.2 \\
BERT & 64.5 & 20.8 & 29.5 & 18.1 \\
ELMo \scsum & 65.4 & 46.2 & 72.7 & 59.3 \\
ELMo \scavg & 56.2 & -9.4 & -0.9 & -22.5 \\
\hline
\multicolumn{5}{c}{Our work} \\\hline
\wa & 54.6 & -10.6 & -32.3 & -27.2 \\
\ws & 75.8 & 57.9 & 75.4 & 60.0 \\ 
\la & 54.6 & -14.8 & -41.1 & -14.8 \\
\sro & 55.5 & 3.2 & 2.2 & 4.1 \\
\wlo & \underline{77.4} & \underline{60.5} & \underline{76.6} & \underline{61.9}

\end{tabular}
\caption{Sentence specificity results on test sets from four domains (accuracy (\%) for News and Spearman correlations (\%) for others). Highest numbers for the models described in this work are underlined.}%
\vspace{-0.1in}
\label{sent-spe-gen-res}
\end{table}

Our averaging-based models (\wa, \la) failed on this task, either giving the majority class accuracy or negative correlations. So, we also evaluate \ws, which sums word embeddings instead of averaging and 
shows strong performance compared to the other models. 

While the model from \citet{li2015fast} performs quite well in the news domain, its performance drops on other domains, indicating some amount of overfitting. On the other hand, \ws and \wlo, which are trained on a large number of paraphrases, perform consistently across the four domains and both outperform the supervised models on Yelp. Additionally, our %
\wlo model outperforms all our other models, achieving comparable performance to the supervised methods. 

Among pretrained models, BERT, Skip-thought, ELMo \scsum, and GloVe \scsum show slight correlations with specificity, while InferSent performs strongly across domains. %
InferSent uses supervised training on a large manually-annotated dataset (SNLI) 
while \ws and \wlo are trained on automatically-generated paraphrases and still show results comparable to InferSent.

\begin{table}[t]
\setlength{\tabcolsep}{7pt}
\centering
\small
\begin{tabular}{l|c|c}
 & Full & Length norm. \\ 
\hline
Majority baseline & 54.6 & 50.1 \\ \hline
\wa & 54.6 & 69.0 \\
\ws & 75.8 & 68.6\\ 
\la & 54.6 & 69.6 \\
\sro & 55.5 & 67.0 \\
\wlo & \textbf{77.4} & \textbf{70.1} \\

\end{tabular}
\caption{Accuracy (\%) for the %
specificity News test set, in both the original  and length normalized conditions. Highest numbers in each column are in bold.}
\label{sent-equal-len-spe-gen-res}
\end{table}

To control for effects due to 
sentence length, we design another experiment in which sentences from News training and test are grouped by length, and thresholds are tuned on the group of length $k$ and tested on the group of length $k-1$, for all $k$, leading to a pool of 3582 test sentences.
 
Table~\ref{sent-equal-len-spe-gen-res} shows the results. In this length-normalized experiment, the averaging models demonstrate much better performance and even outperform \ws, but still \wlo has the best performance.

\section{Analysis}

\begin{table}[t]
\setlength{\tabcolsep}{5pt}
\centering
\small
\begin{tabular}{l|ccc}
        & Entailment & Neutral & Contradiction \\
\hline
GloVe & 42.5 & 53.8 & 39.6 \\
InferSent & 78.3 & 57.2 & 55.7 \\
Skip-thought & 62.5 & 54.3 & 57.3 \\
ELMo & 78.3 & 58.3 & 63.4 \\
BERT & 65.0 & 55.7 & 56.3 \\
\hline
\wa & 77.5 & 50.0 & 57.2 \\
\ws & 75.0 & 54.7 & 57.7 \\ 
\la &  71.7 &  49.5 & 52.4 \\
\sro & 65.0 & 49.5 & 48.6 \\
\wlo & 75.8 & 54.7  & 57.2 \\

\end{tabular}
\caption{Percentage of cases in which hypothesis has larger entropy (or smaller norm for non-probabilistic models) than premise for 
equal-length sentence pairs %
in the SNLI test set. In this setting, GloVe and ELMo would give the same results under either \scsum or \scavg.}
\vspace{-0.1in}
\label{snli-res}

\end{table}

\subsection{Sentence Entailment}
\citet{vilnis2014word} explored whether their Gaussian word entropies captured the lexical entailment relationship. Here we analyze the extent to which our representations capture sentential entailment.

We test models on the SNLI test set, assuming that for a given premise $p$ and hypothesis $h$, $p$ is more specific than $h$ for entailing sentence pairs. To avoid effects due to sentence length, we only consider $\langle p, h\rangle$ pairs with the same length. After this filtering, entailment/neural/contradiction categories have 120/192/208 instances respectively. We encode each sentence and calculate the percentage of cases in which the hypothesis has larger entropy (or smaller norm for non-probabilistic models) than the premise. Under an ideal model, this would happen with 100\% of entailing pairs while showing random results (50\%) for the other two types of pairs. 

As shown in Table~\ref{snli-res}, our best  paraphrase-trained models show similar trends to InferSent, achieving around 75\% accuracy in the entailment category and around 50\% accuracy in other categories. Although ELMo can also achieve similar accuracy in the entailment category, it seems to conflate entailment with contradiction, where it shows the highest percentage of 
all models. Other models, including BERT, GloVe, and Skip-thought, are much closer to random (50\%) for entailing pairs.

\subsection{Lexical Analysis}

\begin{table}[t]
\small
\setlength{\tabcolsep}{2pt}
\centering
\begin{tabular}{c|c|c|c}
        \multicolumn{2}{c|}{Small norm} & \multicolumn{2}{c}{Large norm}\\
\hline
        small abs. ent. & small ent. & small abs. ent. & small ent.\\
\hline
, & addressing & staveb & cenelec\\
/ & derived & jerusalem & ohim\\
by & decree & trent & placebo\\
an & fundamental & microwave & hydrocarbons\\
gon & beneficiaries & brussels & iec\\
as & tendency & synthetic & paras\\
having & detect & christians & allah\\
a & reservations & elephants & milan\\
on & remedy & seldon & madrid\\
for & eligibility & burger & $\pm$\\
from & film-coated & experimental & ukraine\\
'd & breach & alison & intravenous\\
--- & exceed & 63 & electromagnetic\\
his & flashing & prophet & 131\\
' & objectives & diego & electrons\\
upon & cue & mallory & northeast\\
under & commonly & \"{o} & blister\\
towards & howling & natalie & http\\
's & vegetable & hornblower & renal\\
with & bursting & korea & asteroid\\
\end{tabular}
\caption{Examples showing top-20 lists of large-norm or small-norm words ranked based on small absolute entropy or small entropy in \wlo.}
\label{lexical-examples-full}
\end{table}

\wlo associates translation and scaling parameters with each word, allowing us to 
analyze the impact of words on sentence representations. We ranked words under several criteria based on their translation parameter norms and single-word sentence entropies. Table~\ref{lexical-examples-full} shows the top 20 words under each criterion.

\begin{table*}[t]
\scriptsize
\setlength{\tabcolsep}{2pt}
\centering
\begin{tabular}{|p{0.27\textwidth}|p{0.19\textwidth}|p{0.27\textwidth}|p{0.24\textwidth}|}
\hline
        \multicolumn{2}{|c|}{\ws} & \multicolumn{2}{c|}{\wlo}\\
\hline
        \multicolumn{1}{|c|}{largest norm (specific)} & \multicolumn{1}{c|}{smallest norm (general)} & \multicolumn{1}{|c|}{smallest entropy (specific)} & \multicolumn{1}{c|}{largest entropy (general)} \\
\hline
this regulation shall not apply to wine grape products, with the exception of wine vinegar, spirit drinks or flavoured wines. & oh, man, you're gonna... you're just gonna get it, vause\textsuperscript{*}, aren't you ? & under a light coating of dew she was a velvet study in reflected mauve with rose overtones against the indigo nightward\textsuperscript{*} sky. & oh, man, you're gonna... you're just gonna get it, vause\textsuperscript{*}, aren't you? \\\hline
operating revenue community subsidies other subsidies/revenue\textsuperscript{*} total (a) operating expenditure staff administration operating activities total (b) operating result (c=a–b) &
okay, i know you don't get relationships, like, at all, but i don't need to screw anyone for an ``a.''  & a similar influenza disease occurred in 47\% of patients who received plegridy 125 micrograms every 2 weeks, and 13\% of the patients were given placebo. & 'authorisation' means an instrument issued in any form by the authorities by which the right to carry on the business of a credit institution is granted;\\
\hline
\end{tabular}
\caption{Examples of most general and specific sentences for selected lengths (* = mapped to unknown symbol). %
}
\vspace{-0.05in}
\label{sentence-examples}
\end{table*}

Words with small norm and small absolute entropy have little effect, both in terms of meaning and specificity; they are mostly function words. 
Words with large norm and small entropy have a large impact on the sentence while also making it more specific. %
They are organization names (\emph{cenelec}) or technical terms found in medical or scientific literature. When they appear in a sentence, they are very likely to appear in its paraphrase. 

Words with large norm and small absolute entropy contribute to the sentence semantics but do not make it more specific. Words like \emph{microwave} and \emph{synthetic} appear in many contexts and have multiple senses. Names (\emph{trent}, \emph{alison}) also appear in many contexts. Words like these often appear in a sentence's paraphrase, but can also appear in many other sentences in different contexts. 

Words with small norm/entropy make sentences more specific 
but do not lend themselves to a precise characterization. They affect sentence meaning, but can be expressed in many ways. 
For example, when \emph{beneficiaries} appears in a sentence, its paraphrase often has a synonym like \emph{beneficiary}, \emph{heirs}, or \emph{grantees}. These words may have multiple senses, but it appears more that they correspond to concepts with many valid ways of expression.

\subsection{Sentential Analysis}

We subsample the ParaNMT training set and group sentences by length. For each model and length, we pick the sentence with either highest/lowest entropy or largest/smallest norm values. Table~\ref{sentence-examples} shows some examples. %
\ws tends to choose conversational sentences as general and those with many rare words as specific. \wlo favors literary and technical/scientific sentences as most specific, and bureaucratic/official language as most general.

\begin{table}[t]
\setlength{\tabcolsep}{5pt}
\centering
\small
\begin{tabular}{l|cc|cc}
 &\multicolumn{2}{c|}{With Prior} &\multicolumn{2}{c}{Without Prior} \\ 
        & Acc. & $F_{1}$ & Acc. & $F_{1}$  \\
\hline
\wlo & 77.4 & 78.4  & 67.9 & 68.2 \\

\end{tabular}
\caption{Accuracy (\%) and $F_{1}$ score (\%) for specificity News test set with and without prior regularization.}
\label{prior-res}
\end{table}

\subsection{Effect of Prior Regularization}
As shown in Table~\ref{prior-res}, there is a large performance improvement after adding prior regularization for avoiding degenerate solutions.

\begin{table}[t]
\setlength{\tabcolsep}{5pt}
\centering
\small
\begin{tabular}{l|c}
        & STS Benchmark\\
\hline
\wa & 73.4 \\
\la & 73.6 \\
\sro & \textbf{74.3} \\
\wlo & 73.7 \\

\end{tabular}
\caption{Pearson correlation (\%) for STS benchmark test set. Highest number is in bold.}
\label{sts-res}
\end{table}

\subsection{Semantic Textual Similarity}
Although semantic textual similarity is not our target task, we still include the performance of our models on the STS benchmark test set in Table~\ref{sts-res} to show that our models are competitive with standard strong baselines. 
When using probabilistic models to predict sentence similarity during test time, we let $v_1=\mathit{concat}(\mu_1,\Sigma_1)$, $v_2=\mathit{concat}(\mu_2,\Sigma_2)$, where $\mathit{concat}$ is a concatenation operation, and predict sentence similarity via $\mathit{cosine}(v_1, v_2)$, since we find it performs better than solely using the mean vectors. 
The two probabilistic models, \sro and \wlo, are able to outperform the baselines slightly.

\section{Related Work}

Our models are related to work in learning probabilistic word embeddings~\cite{vilnis2014word,ath2017multimodal,athiwaratkun-etal-2018-probabilistic} and text-based 
VAEs~\cite[\emph{inter alia}]{Miao2016neural,bowman16gen,pmlr-v70-yang17d,pmlr-v80-kim18e,xu2018spherical}. 
The \wlo is also related to flow-based VAEs~\cite{rezende15norm-flow,kingma2016improved}, where hidden layers are 
viewed 
as operators over the density function of latent variables. 

Previous work on sentence specificity relies on hand-crafted features or direct training on annotated data~\cite{louis2011auto,li2015fast}. Recently, \citet{ko2018domain} used domain adaptation for this problem when only the source domain has annotations. 
Our work also relates to learning sentence embeddings from paraphrase pairs~\cite{wieting-16-full,para-nmt-acl-18}.

\section{Conclusion}
We trained sentence models on paraphrase pairs and showed that they naturally capture specificity and entailment.  
Our proposed \wlo model, which treats each word as a linear transformation operator, achieves the best performance and lends itself to analysis.

\section*{Acknowledgments}
We would like to thank the anonymous reviewers, NVIDIA for donating GPUs used in this research, Jessy Li for clarifying the experimental setup used in~\citet{li2015fast}, and Google for a faculty research award to K.~Gimpel that partially supported this research.

\bibliography{acl2020}
\bibliographystyle{acl_natbib}
\appendix

\section{Supplementary Material}

\subsection{Hyperparameters}
For all experiments, the dimension of word embeddings and word operator is 50. The dimension of LSTM is 100. The dimension of Gaussian distribution for \sro is 100. Mini-batch size is 100. For LSTM, \sro, and \wlo, we scramble training sentences with a probability of 0.4. For baseline models, the margin $\delta$ is $0.4$. For other models, $\delta$ is 1. All models are randomly initialized and trained with Adam~\cite{kingma2014adam} using learning rate of 0.001.

\end{document}